# Measuring What Matters: Intrinsic Distance Preservation as a Robust Metric for Embedding Quality

Steven N. Hart, PhD* and Thomas Tavolara, Ph.D.

*Department of Laboratory Medicine and Pathology, Mayo Clinic, 200 1$^{st}$ St. SW*
*Rochester, MN 55901, USA*
*Email: hart.steven@mayo.edu*

*Corresponding Author

## Abstract

Unsupervised embeddings are fundamental to numerous machine learning applications, yet their evaluation remains a challenging task. Traditional assessment methods often rely on extrinsic variables, such as performance in downstream tasks, which can introduce confounding factors and mask the true quality of embeddings. This paper introduces the Intrinsic Distance Preservation Evaluation (IDPE) method, a novel approach for assessing embedding quality based on the preservation of Mahalanobis distances between data points in the original and embedded spaces. We demonstrate the limitations of extrinsic evaluation methods through a simple example, highlighting how they can lead to misleading conclusions about embedding quality. IDPE addresses these issues by providing a task-independent measure of how well embeddings preserve the intrinsic structure of the original data. Our method leverages efficient similarity search techniques to make it applicable to large-scale datasets. We compare IDPE with established intrinsic metrics like trustworthiness and continuity, as well as extrinsic metrics such as Average Rank and Mean Reciprocal Rank. Our results show that IDPE offers a more comprehensive and reliable assessment of embedding quality across various scenarios. We evaluate PCA and t-SNE embeddings using IDPE, revealing insights into their performance that are not captured by traditional metrics. This work contributes to the field by providing a robust, efficient, and interpretable method for embedding evaluation. IDPE's focus on intrinsic properties offers a valuable tool for researchers and practitioners seeking to develop and assess high-quality embeddings for diverse machine learning applications.

# Introduction

Unsupervised embeddings are fundamental in modern machine learning and data analysis, underpinning various downstream tasks. However, evaluating the quality of these embeddings poses significant challenges. Traditionally, embedding quality has been assessed through extrinsic methods, such as performance in classification, clustering, or ranking tasks[1], [2]. Others have considered this question [3], using a series of different analytical perspectives including α-Req [4], RankMe [5], [6], and NeSum [7]. These methods found that statistical properties of data—independent of labels—could serve as surrogates for unsupervised learning (e.g., coherence and stable rank) when assessing performance in a downstream classification task.

While these approaches provide some insights into embedding quality without labels, they expose a fundamental flaw: the evaluation becomes a hybrid measure of embedding quality and task complexity. This issue violates the "controlled variable principle" or *ceteris paribus* (Latin for "all other things being equal"), which states that to determine the effect of one independent variable on a dependent variable accurately, all other influencing factors must remain constant.

To illustrate this flaw, consider a dataset of two-dimensional points clustered into two distinct groups. We generate embeddings using three different algorithms: Principal Component Analysis (PCA) [8], t-Distributed Stochastic Neighbor Embedding (t-SNE) [9], and Gaussian Random Projection (GRP)[1]. These algorithms were chosen for their contrasting approaches to dimensionality reduction. PCA, a linear method, preserves global structure by maximizing variance. t-SNE, a non-linear method, emphasizes local relationships. GRP offers a fast and scalable approach with approximate distance preservation, suitable for very high-dimensional data but less interpretability and potential variability due to randomness.

We evaluate these embeddings using a downstream classification task where a classifier separates the two clusters based on the embeddings. Suppose PCA maintains the global structure but may not optimally separate the clusters. t-SNE might produce clearer cluster separation by emphasizing local relationships, potentially distorting global relationships. GRP, using random projections, may result in variable preservation of local and global structures. One might intuitively expect t-SNE to perform better in the classification task due to its focus on local structure. PCA might not separate clusters as clearly, and GRP could yield varying results depending on the randomness of the projections. However, the classifier's robustness allows it to achieve similar classification accuracy across all three embeddings, masking the fundamental differences in how these methods preserve the original data structure. This extrinsic evaluation suggests that PCA, t-SNE, and GRP produce embeddings of similar quality, despite their differing preservation of data relationships.

This example highlights a critical problem with extrinsic evaluations: they conflate embedding quality with downstream task performance. The classifier's ability to adapt to different embedding characteristics can lead to misleading conclusions about their true quality. Moreover, this approach fails to capture the distinct strengths and weaknesses of each method. Our work addresses this issue by developing an Intrinsic Distance Preservation Evaluation (IDPE) method that effectively captures and quantifies differences in embedding quality, independent of specific downstream tasks. This approach provides a more nuanced and accurate assessment of embedding algorithms, particularly when methods produce very different yet equally effective embeddings from an extrinsic perspective.

We also present a comparative analysis of IDPE with established intrinsic metrics such as trustworthiness and continuity [10], as well as extrinsic metrics like Average Rank (AR) and Mean Reciprocal Rank (MRR) [11]. Through this analysis, we demonstrate the advantages and limitations of various evaluation approaches, highlight the value of intrinsic evaluation methods in providing more reliable and generalizable measures of embedding quality, and offer insights into selecting appropriate evaluation metrics for different embedding scenarios. By advancing the use of intrinsic evaluation methods, we seek to improve the assessment of unsupervised embeddings, ultimately contributing to the development of more robust and versatile embedding techniques for a wide range of machine learning applications.

**Methods**

**Dataset generation:** We used the make_blobs function from the sklearn.datasets module to create synthetic data [12]. This function generates isotropic Gaussian blobs for clustering, providing a clear and controlled environment to test the embedding algorithms. The data consisted of 500 samples, evenly distributed across two distinct clusters. The clusters were separated well to ensure that the inherent structure was clear and distinguishable. To simulate different noise conditions, we varied the standard deviation of the Gaussian blobs, using noise levels of 0.6, 1.0, and 1.5. Increasing the noise level made the clusters more overlapping and less distinct, thereby challenging the embedding algorithms to maintain the data structure under these conditions. We also explored the impact of three different dimensionalities: 2, 128, and 512.

Secondarily, we assessed the performance of each metric on four additional synthetic datasets with distinct characteristics: circles, moons, s-curve, and swiss roll. The datasets were generated using the make_moons, make_s_curve, make_swiss_roll, and make_circles functions from the sklearn.datasets module. For each dataset, we generated 1000 samples with different levels of noise (0.1, 1, 2, 5, 10). The generated datasets were then standardized using StandardScaler. The process was repeated 10 times to evaluate the reproducibility and measurement error.

**Average Rank (AR)**: The Average Rank (AR) metric measures the average position (rank) of relevant items in a ranked list generated by an embedding or retrieval system. For each point $i$ in the dataset, the rank of its $k$-nearest neighbors in the original space is compared to their rank in the embedding space. The rank is averaged over all points to get the AR. Lower AR indicates better neighborhood preservation.

Let $q$ be a query and $R(q)$ be the ranked list of results for query $q$. Let $G(q)$ be the set of ground truth relevant items for query $q$. The rank of item $g \in G(q)$ in the ranked list $R(q)$ is denoted as $rank(g)$.

The formula for AR for a single query $q$ is given by:

$$AR(q) = \frac{1}{|G(q)|} \sum_{g \in G(q)} rank(g)$$

where $|G(q)|$ is the number of relevant items for query $q$.

To compute the overall AR across all queries in a dataset, we average the AR values for all queries. Let $Q$ be the set of all queries. The overall AR is given by:

$$AR(q) = \frac{1}{|Q|} \sum_{q \in Q} \left( \frac{1}{|G(q)|} \sum_{g \in G(q)} rank(g) \right)$$

This formula calculates the average rank of the relevant items for each query and then averages these ranks across all queries.

**Average Normalized Rank (ANR)**: This metric normalizes the rank by the maximum possible rank to make it comparable across datasets of different sizes. It is computed as:

$$ANR = \frac{1}{N} \sum_{i=1}^{N} \frac{1}{k} \sum_{j=1}^{k} \frac{rank(j)}{N}$$

where $N$ is the total number of points and $rank(j)$ is the rank of the $j$-th neighbor of point $i$ in the embedding space.

**Mean Reciprocal Rank (MRR)**: This metric evaluates the rank of the nearest neighbors in the embedding space for each point in the original space. It is the average of the reciprocal ranks of the true nearest neighbors. Higher MRR indicates better neighborhood preservation.

$$MRR = \frac{1}{N} \sum_{i=1}^{N} \frac{1}{rank_{true}(i)}$$

where $rank_{true}(i)$ is the rank of the true nearest neighbor in the embedding space for point *i*.

**Trustworthiness (T):** This metric measures how much the neighborhood relationships in the embedding space can be trusted compared to the original space. It is defined as:

$$T(k) = 1 - \frac{2}{Nk(2N - 3k - 1)} \sum_{i=1}^{N} \sum_{j \in U_i^k} (rank(i,j) - k)$$

Where $U_i^k$ is the set of points that are in the *k*-nearest neighbors of *i* in the embedding space but not in the original space, and $rank(i,j)$ is the rank of point *j* in the original space. High trustworthiness means that points that are close in the embedding space were also close in the original space, indicating that the embedding space is reliable.

**Continuity (C):** This metric evaluates how well the neighborhood relationships in the original space are preserved in the embedding space. It is defined as:

$$C(k) = 1 - \frac{2}{Nk(2N - 3k - 1)} \sum_{i=1}^{N} \sum_{j \in U_i^k} (rank(i,j) - k)$$

Where $V_k^i$ is the set of points that are in the *k*-nearest neighbors of *i* in the original space. High continuity means that points that were close in the original space remain close in the embedding space, showing that the embedding space accurately reflects the original neighborhood structure.

### Intrinsic Distance Preservation Evaluation (IDPE)

Intrinsic Distance Preservation Evaluation (IDPE) is a novel approach for evaluating the quality of unsupervised embeddings by focusing on intrinsic metrics rather than relying on downstream tasks. This method aims to assess how well embeddings preserve the original distances between data points in high-dimensional space, avoiding the introduction of confounding variables associated with task-specific evaluations. To do this, IDPE assesses the quality of embeddings by measuring how well the embeddings preserve the original distances between data points in the high-dimensional space. The fundamental idea is to compare the distances between vectors in the original data (truth set) and their corresponding vectors in the embedding space. Smaller distances mean more similarity to the original feature space. This concept is illustrated in **Figure 1**.

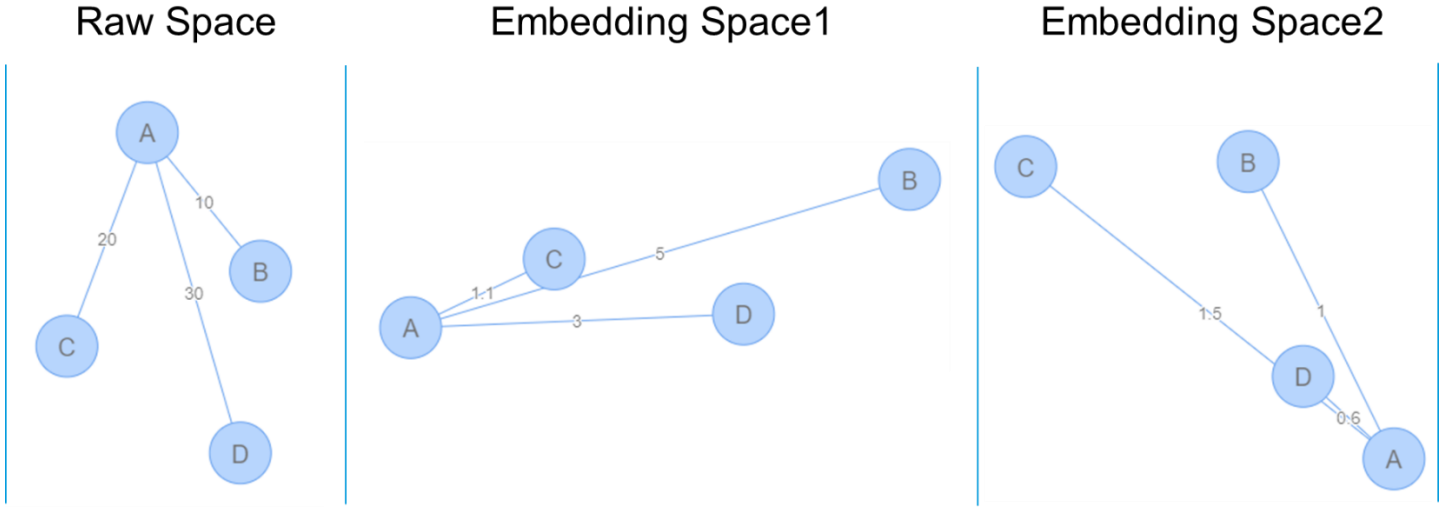

*Figure 1. Conceptual diagram of IDPE.* In raw space, the Mahalanobis distance between A, B, and C are 10, 20, and 30, respectively. The top 2 nearest neighbors to A are nodes B and C. This represents the truth set. In Embedding space 1, the two nearest neighbors to A are C & D, and in Embedding space 2, B and D are nearest to A. For IDPE, the score would be 30+20 (C+D)=50 for Embedding space 1 and 30+10=40 (B+D) for Embedding space 2. So, embedding Space 2 would have a lower score, hence would be a better embedding.

Let $X = \{x_1, x_2, \ldots, x_n\}$ be the set of original high-dimensional data points. Let $Z = \{z_1, z_2, \ldots, z_n\}$ be the set of corresponding embeddings in a lower-dimensional space. Let $d_m(a, b)$ denote the Mahalanobis distance between two vectors $a$ and $b$ such that $\sqrt{(a-b)^T S^{-1}(a-b)}$ where $S$ is the covariance matrix of the original data $X$. Let $\mathbb{N}(x_i)$ denote the set of $k$-nearest neighbors of $x_i$ in the original space. Let $\mathbb{N}(z_i)$ denote the set of $k$-nearest neighbors of $z_i$ in the embedding space.

Mathematically, the steps can be formulated as follows. For each $x_i \in X$, calculate:

$$D_x(x_i) = \{d_m(x_i, x_j) \mid x_j \in \mathbb{N}(x_i)\}$$

Similarly, for each $z_i \in Z$, calculate:

$$D_z(z_i) = \{d_m(z_i, z_j) \mid z_j \in \mathbb{N}(z_i)\}$$

For each data point $x_i$, compute the preservation error as:

$$Error(x_i, x_j) = |D_x(x_i, x_j) - D_z(z_i, z_j)|$$

Aggregate the preservation errors for all data points to obtain a global score:

$$IDPE = \frac{1}{n} \sum_{i=1}^{n} \sum_{j \in N(x_i)} |D_x(x_i, x_j) - D_z(z_i, z_j)|$$

**Implementation details for IDPE**

This section outlines the logic and implementation details for calculating IDPE using FAISS[13], an efficient library for similarity search and clustering of dense vectors. The only other necessary libraries are scikit-learn and NumPy for calculating mean squared error [12]. The full code is shown in **Box 1**.

We start with two datasets: `original_data`, containing the original high-dimensional data, and

> **Box 1. The IDPE Algorithm**
>
> ```python
> import faiss
> import numpy as np
> from sklearn.metrics import pairwise_distances, mean_squared_error
>
> def idpe(original_data, embedded_data, k=5):
>
>     def _build_index(train_df):
>         d = train_df.shape[1]
>         IndexFlatL2_index = faiss.IndexFlatL2(d)
>         IndexFlatL2_index.add(train_df)
>         return IndexFlatL2_index
>
>     def get_mahalanobis_distance(vector_1, vector_2, inv_cov_matrix):
>         distance = pairwise_distances(vector_1, vector_2, metric='mahalanobis', VI=inv_cov_matrix)
>         return distance[0][0]
>
>     def get_distances(index_to_search=None, truth_index=None, inv_cov_matrix=None):
>         distances = []
>         for compressed_index in range(index_to_search.shape[0]):
>             raw_indexes = list(index_to_search[compressed_index])
>             for i in raw_indexes:
>                 vector_1 = truth_index.reconstruct(compressed_index).reshape(1, -1)
>                 vector_2 = truth_index.reconstruct(int(i)).reshape(1, -1)
>                 distances.append(get_mahalanobis_distance(vector_1, vector_2, inv_cov_matrix))
>         return distances
>
>     cov_matrix = np.cov(original_data, rowvar=False)
>     inv_cov_matrix = np.linalg.inv(cov_matrix)
>
>     truth_index = _build_index(original_data)
>     test_index = _build_index(embedded_data)
>     true_distances, true_indexes = truth_index.search(original_data, k)
>     test_distances, test_indexes = test_index.search(embedded_data, k)
>     calculated_distances = get_distances(index_to_search=test_indexes, truth_index=truth_index,
>                                          inv_cov_matrix=inv_cov_matrix)
>     mse_true = mean_squared_error(true_distances.flatten(), calculated_distances)
>     return mse_true
> ```

embedded_data, containing the low-dimensional embeddings generated by the method under evaluation, such as PCA or t-SNE. The first step in implementing IDPE is to build an index for the raw data using FAISS. This involves several key steps. First, we define the number of nearest neighbors (k) to consider, which is set to 5. This ensures that we are capturing local neighborhood information crucial for assessing how well the embedding preserves these relationships.

Next, we build an index for the original data and the embedded data using FAISS. We create the FAISS index with the function `_build_index`, which initializes the index based on the dimensionality of the data. This dimensionality is essential for initializing the FAISS index, as it defines the space in which we will search for nearest neighbors. The original data is then added to the FAISS index with `truth_index.add(original_data)` and similarly, the embedded data with `test_index.add(embedded_data)`. These steps populate the indices with our data points, making them ready for nearest neighbor searches.

Once the indexes are built, the top k nearest neighbors for each data point are calculated in both the original and embedded datasets. For the original data, `true_distances, true_indexes = truth_index.search(original_data, k)` is used to search the FAISS index for the nearest neighbors of each data point. This returns the distances and indexes of the nearest neighbors, capturing the local neighborhood relationships in the high-dimensional space. Similarly, for the embedded data, `test_distances, test_indexes = test_index.search(embedded_data, k)` performs the nearest neighbor search in the embedded space.

To quantify the preservation of distances, we need to calculate the differences between the original and embedded distances for the nearest neighbors. We define a helper function `get_mahalanobis_distance` to compute the Mahalanobis distance between two vectors using the inverse covariance matrix. Another helper function, `get_distances`, iterates over the nearest neighbors and calculates the Mahalanobis distances between the original data points and their nearest neighbors as determined in the embedded space. Finally, the mean squared error between the true distances in the original space and the calculated distances based on the embeddings is computed to quantify how well the embedding preserves the original distances. Lower values indicate better preservation of the original data structure.

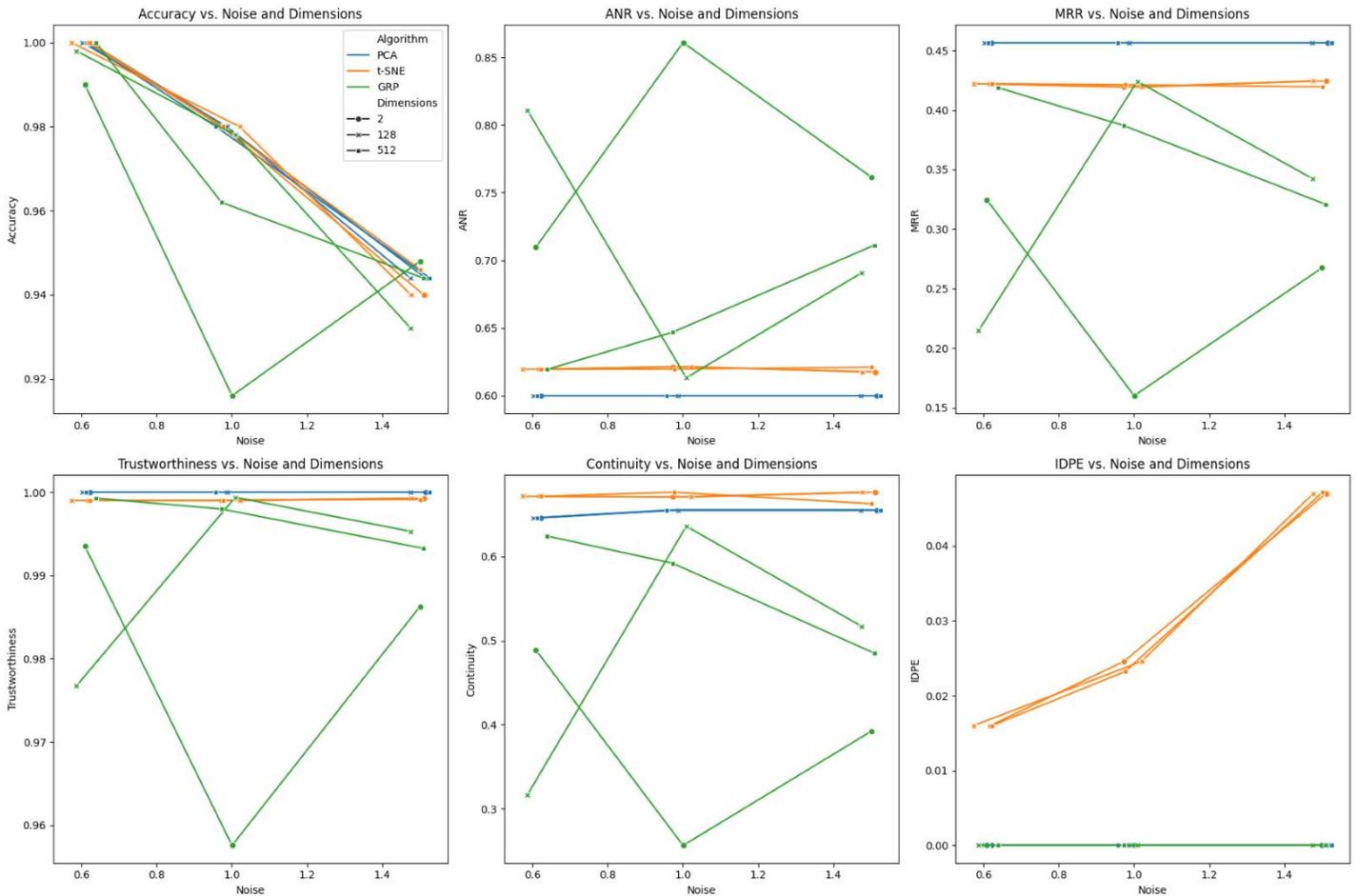

*Figure 2. Evaluation of PCA, t-SNE, and Gaussian Random Projection (GRP) embeddings across various noise levels and dimensions.* The metrics assessed include Accuracy from logistic regression classification, Average Normalized Rank (ANR), Mean Reciprocal Rank (MRR), Trustworthiness, Continuity, and Intrinsic Distance Preservation Evaluation (IDPE).

## *Results*

To evaluate the performance of PCA and t-SNE embeddings, we generated synthetic datasets with varying noise levels and dimensionalities **(Figure 2)**. This systematic approach allowed us to investigate the impact of different data characteristics on the quality of embeddings produced by these algorithms. Given that PCA preserves the original structure while t-SNE distorts it due to its non-linear nature, we interpreted the results within these contexts. Specifically, we assessed how well each embedding method preserved both local and global data structures using a range of intrinsic metrics, including AR, MRR, Trustworthiness, Continuity, and the newly introduced Intrinsic Distance Preservation Evaluation (IDPE).

As noise levels increase, we anticipated that PCA would maintain more stable accuracy and trustworthiness scores compared to t-SNE, which might show more variation due to its sensitivity to local structures. Similarly, the local neighborhood metrics (AR and MRR) should consistently favor t-SNE across all conditions, demonstrating its strength in capturing local relationships. To determine whether the IDPE effectively evaluates the preservation of distances in embeddings, we compared its results with the other intrinsic metrics to identify consistent and supporting evidence.

As shown in **Figure 2**, an increase in noise on simulated blobs results in the expected decrease in logistic regression classification accuracy across all transformations (PCA, t-SNE, and GRP). However, GRP's accuracy varies significantly, especially in lower dimensions, and is generally less stable across metrics such as ANR, MRR, Trustworthiness, and Continuity. This instability is likely due to the randomization inherent in GRP.

MRR and ANR, which reflect global structure, show better scores for PCA-transformed data compared to t-

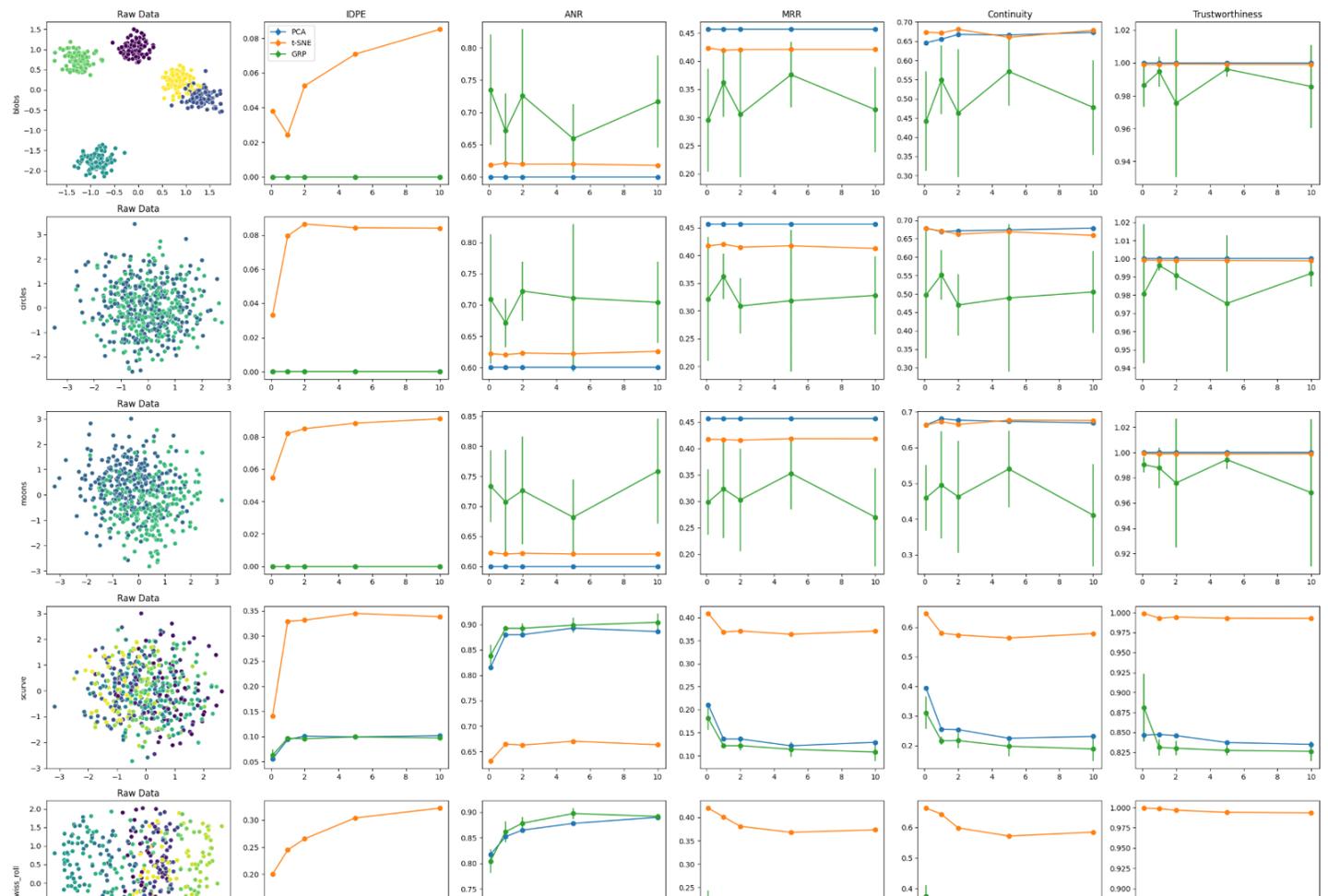

*Figure 3. Effect of Noise on Metric Performance by Transformation and Dataset with Error Bars.* Each row represents different synthetic datasets generated with varying noise levels and dimensionalities. The columns represent different metrics used to evaluate the embeddings, including IDPE, ANR, MRR, Continuity, and Trustworthiness.

SNE, indicating PCA's strength in preserving global data structures. Conversely, Continuity, which emphasizes local structure, has higher scores for t-SNE-transformed data, validating t-SNE's effectiveness in preserving local relationships.

PCA consistently exhibits low IDPE values, confirming its superior performance in preserving intrinsic distances and maintaining global data structures. Despite higher IDPE values indicating distortions in global distances, t-SNE achieves similar classification accuracy to PCA, highlighting the robustness of local structure preservation in classification tasks.

We applied PCA, t-SNE, and GRP to understand the relationship between other data structures. Noise levels were assessed at multiple intervals (0.1, 1, 2, 5, 10), and experiments were repeated 10 times. **Figure 3** presents a comprehensive analysis of how noise affects various metrics for three dimensionality reduction techniques: PCA, t-SNE, and GRP. Each row represents a different dataset (blobs, circles, moons, s-curve, and swiss roll), and the columns show the raw data followed by the metrics: IDPE, ANR, MRR, Continuity, and Trustworthiness.

IDPE consistently highlights PCA's superior performance in preserving intrinsic distances, as evidenced by its low values across all noise levels and datasets. In contrast, t-SNE shows higher IDPE values that increase with noise, reflecting its focus on preserving local structures at the expense of global distance preservation. GRP exhibits less variability in IDPE scores compared to other metrics, indicating a somewhat stable performance in maintaining distances despite its inherent randomness.

This relative stability in GRP's IDPE scores, despite its variability in other metrics, can be attributed to how GRP and IDPE interact. GRP, by design, aims to approximately preserve pairwise distances through random projections, which can result in a consistent average performance across different runs. However, this consistency in distance preservation does not necessarily translate to other structural aspects of the data, which are captured by metrics like Continuity, Trustworthiness, ANR, and MRR. These metrics assess local and global structural preservation in different ways, often revealing more pronounced variability in GRP's performance.

## Discussion

Evaluating the quality of embeddings is crucial in various fields, including dimensionality reduction, manifold learning, and information retrieval. Traditional metrics often focus on either local or global relationships or rely on task-specific evaluations. In this context, we introduce the IDPE metric and compare it with established measures.

IDPE offers a simple yet comprehensive approach to assessing embedding quality. By leveraging FAISS for efficient computation, it provides a balanced measure encompassing both local and global relationships within the data structure. The primary advantage of IDPE lies in its directness and ability to evaluate embeddings holistically without the confounding influence of external tasks.

Established metrics like Trustworthiness and Continuity offer robust theoretical foundations in dimensionality reduction and manifold learning. Trustworthiness assesses the preservation of local neighborhood structure, while Continuity evaluates the correspondence of neighbors between original and embedded spaces. Despite their strengths, these metrics involve significant computational complexity, particularly for large datasets, and primarily focus on local structures.

In the realm of information retrieval and recommendation systems, MRR is widely used to evaluate ranking quality. It measures the rank of the first relevant item for each query, offering intuitive interpretation. However, MRR depends on predefined relevance judgments and is sensitive to the top-ranked item's position.

Other metrics such as Average Rank (AR and ANR are valuable for evaluating embedding performance in ranking tasks. However, IDPE provides advantages in certain applications by offering a direct, intrinsic assessment of embedding quality. It avoids confounding factors associated with task-specific evaluations and focuses on preserving intrinsic geometric relationships, making it particularly valuable for tasks where maintaining the original data structure is crucial.

Our evaluation of PCA and t-SNE embeddings using IDPE revealed interesting findings. IDPE values remained consistent across noise levels and were similar between PCA and t-SNE, suggesting that both methods effectively balance global and local structure preservation. This contrasts with traditional metrics, where PCA and t-SNE often show distinct performance differences.

When comparing IDPE with Trustworthiness and Continuity, it's important to consider their specific goals and computational complexities. While Trustworthiness and Continuity offer robust theoretical foundations, they involve significant computational complexity for large datasets. IDPE, leveraging efficient nearest neighbor search implementations like FAISS, provides a computationally efficient alternative while capturing both local and global distance preservation.

The choice between IDPE, AR, ANR, Trustworthiness, Continuity, and MRR depends on specific application requirements, including the need for intrinsic versus task-specific evaluation, computational efficiency, and the importance of preserving local versus global data structures. Our findings support the use of IDPE alongside established metrics, providing a comprehensive assessment of embedding quality.

In conclusion, IDPE offers a valuable tool for evaluating intrinsic distance preservation in various embedding methods. Its balanced approach to capturing both local and global relationships, combined with computational efficiency, makes it a promising complement to existing evaluation metrics in the field of embedding analysis.